\documentclass[11pt,a4paper]{article}

\makeatletter
\newcommand{\@BIBLABEL}{\@emptybiblabel}
\newcommand{\@emptybiblabel}[1]{}
\makeatother

\usepackage[hyperref]{naaclhlt2018}
\usepackage{times}
\usepackage{latexsym}

\usepackage{url}

\usepackage{xspace}
\usepackage{graphicx}
\usepackage{mathrsfs, amsmath, amssymb}
\usepackage{scrextend}
\usepackage{booktabs}
\usepackage{paralist}
\usepackage[tight-spacing=true]{siunitx}
\usepackage{etoolbox,multirow,adjustbox}
\usepackage{color,soul}
\usepackage{xcolor}
\usepackage{arydshln}
\usepackage{relsize}
\robustify\bfseries

\usepackage{tikz}
\usetikzlibrary{positioning}
\usetikzlibrary{calc}
\usetikzlibrary{fit}
\usetikzlibrary{decorations.text}
\usetikzlibrary{decorations.pathreplacing}
\usetikzlibrary{shapes.misc}
\usetikzlibrary{shapes.geometric,arrows}

\aclfinalcopy 
\def\aclpaperid{1051} 

\newcommand{\dataset}[1]{\textsf{#1}\xspace}
\newcommand{\resource}[2][]{\textsc{#2}#1\xspace}
\newcommand{\CNN}{\ensuremath{\operatorname{CNN}}\xspace}

\newcommand{\cond}{\resource{Cond}}
\newcommand{\gen}{\resource{Gen}}
\newcommand{\SP}{\cond}
\newcommand{\CP}{\gen}

\newcommand{\tabref}[1]{Table~\ref{#1}\xspace}
\newcommand{\figref}[1]{Figure~\ref{#1}\xspace}
\newcommand{\secref}[1]{\S\ref{#1}\xspace}

\newcommand\loss{\mathcal{L}}
\newcommand\xent{\mathcal{X}}
\newcommand\hmat{\mathbf{H}}
\newcommand\hvec{\mathbf{h}}
\newcommand\dvec{\mathbf{d}}
\newcommand\yvec{\mathbf{y}}
\newcommand\xvec{\mathbf{x}}

\usepackage{subcaption}
\DeclareCaptionFont{10pt}{\fontsize{10pt}{12pt}\selectfont}
\title{What's in a Domain? \\ Learning Domain-Robust Text Representations \\ using Adversarial Training}

\author{Yitong Li \and Timothy Baldwin \and Trevor Cohn \\
  School of Computing and Information Systems\\
  The University of Melbourne, Australia \\
  {\smaller {\tt yitongl4@student.unimelb.edu.au, \{tbaldwin,tcohn\}@unimelb.edu.au}}
  }

\date{}

\begin{document}

\maketitle
\begin{abstract}
  Most real world language problems require learning from heterogenous corpora,
  raising the problem of learning robust models which generalise well to both similar (\emph{in domain}) and dissimilar (\emph{out of domain}) instances to those seen in training.
  This requires learning an underlying task, while not learning irrelevant signals and biases specific to individual domains.
  We propose a novel method to optimise both in- and out-of-domain accuracy based on joint learning of a structured neural model with domain-specific and domain-general components, coupled with adversarial training for domain.
  Evaluating on multi-domain language identification and multi-domain sentiment analysis, we show substantial improvements over standard domain adaptation techniques, and domain-adversarial training.

\end{abstract}

\section{Introduction}

Heterogeneity is pervasive in NLP, arising from corpora being constructed from different sources, featuring different topics, register, writing style, etc. 
An important, yet elusive, goal is to produce NLP tools that are capable of handling all types of texts, such that we can have, e.g., text classifiers that work well on texts from newswire to wikis to micro-blogs.
A key roadblock is application to new domains, unseen in training.
Accordingly, training needs to be robust to domain variation, such that domain-general concepts are learned in preference to domain-specific phenomena, which will not transfer well to out-of-domain evaluation. 
To illustrate, \newcite{bitvai15acl} report learning formatting quirks of specific reviewers in a review text regression task, which are unlikely to prove useful on other texts.

This classic problem in NLP has been tackled under the guise of ``domain adaptation'', also known as unsupervised transfer learning, using feature-based methods to support knowledge transfer over multiple domains~\cite{blitzer2007biographies,daume2007frustratingly,Joshi2012multi,williams2013multi,C16-1038}.
More recently, \newcite{ganin2015unsupervised}  proposed a method to encourage domain-general text representations, which transfer better to new domains.

Inspired by the above methods, in this paper we propose a novel technique for multitask learning of domain-general representations.\footnote{Code, data and evaluation scripts available at \url{https://github.com/lrank/Domain_Robust_Text_Representation.git}}
Specifically, we propose deep learning architectures for multi-domain learning, featuring a \emph{shared} representation, and domain \emph{private} representation.
Our approach generalises the feature augmentation method of \newcite{daume2007frustratingly} to convolutional neural networks, as part of a larger deep learning architecture.
Additionally, we use adversarial training such that the \emph{shared} representation is explicitly discouraged from learning domain identifying information \cite{ganin2015unsupervised}.
We present two architectures which differ in whether domain is conditioned on or generated, and in terms of parameter sharing in forming private representations.

We primarily evaluate on the task of language identification (``LangID'': \newcite{cavnar1994n}),  using the corpora of \newcite{lui2012langid}, which combine large training sets over a diverse range of text domains.
Domain adaptation is an important problem for this task \cite{vrl2014accurate,DBLP:conf/acl/JurgensTJ17}, where text resources are collected from numerous sources, and exhibit a wide variety of language use.  
We show that while domain adversarial training overall improves over baselines, gains are modest. The same applies to twin shared/private architectures, but when the two methods are combined, we observe substantial improvements. Overall, our methods outperform the state-of-the-art \cite{lui2012langid} in terms of out-of-domain accuracy.
As a secondary evaluation, we use the Multi-Domain Sentiment Dataset \cite{blitzer2007biographies}, where we once again observe a clear advantage for our approaches, illustrating  the potential of our technique more broadly in NLP.

\section{Multi-domain Learning}


A primary consideration when formulating models of multi-domain data is how best to use the domain. 
Basic methods might learn several separate models, or simply ignore the domain and learn a single model. 
Neither method is ideal: the former fails to share statistics between the models to capture the general concept, while the latter discards information that can aid classification, e.g.,  domain-specific vocabulary or class skew.

\begin{figure}[t]
  \centering
  \begin{subfigure}[b]{\columnwidth}
    \scalebox{1}{\begin{tikzpicture}[line width=0.02cm]

    \node[align=center,minimum size=0.3cm] at (-3.3, 0) (A2) {$\mathbf{x}_i$};

    \node[align=center,minimum size=0.5cm] at (-1.3, 0.25) (B2) {\small $\CNN_{d_i}^p (\theta^p_{d_i})$};

    \node[draw=none,align=center,minimum size=0.5cm] at (-1.3, 1.2) {\small $\CNN^{s} (\theta^s)$};

    \filldraw[fill=green!20!white, draw=green!20!white,rounded corners] (-0.3, -0.3)rectangle(0.3, 0.3);
    \node[draw=none,align=center,minimum size=0.5cm] at (0.0, +0.0) {\small $\mathbf{h}^p_i$};
    \filldraw[fill=yellow!20!white, draw=yellow!20!white,rounded corners] (-0.3, +0.7)rectangle(0.3, 1.3);
    \node[draw=none,align=center,minimum size=0.5cm] at (0.0, +1.0) {\small $\mathbf{h}^s_i$};

    \node[draw=none,align=center,text width=6cm] at (1.9,+0.0) {$y_i$};

    \draw[->] (-2.8,  0.0)--( -0.35, -0.0);
    \draw[->,rounded corners] (-2.8, 0.0)--(-2.5, 0.0)--(-2.5,1.)--(-0.35,1.);
    \draw[draw=blue,->] ( 0.35, 0.0)--(1.5, 0.0);
    \draw[draw=blue,->,rounded corners] ( 0.35,1.0)--( 1.0, 1.0)--( 1.0, 0.0)--( 1.5, 0.0);
    
    \draw[draw=red!60!white,dashed,->,rounded corners] (0., 1.3)--(0., 1.7)--(1.5, 1.7);

    \node[align=center,minimum size=0.5cm] at (0.9, 1.9) {\small $\operatorname{D}^s(\theta^d)$};
    \node[align=center,minimum size=0.5cm] at (1.8, 1.7) {$d_i$};

    \node[align=center,minimum size=0.5cm] at (0.8, 0.3) {\small $\theta^c$};

\end{tikzpicture}

}
    \caption{Domain-conditional (\cond) model.}
    \label{fig:sp}
  \end{subfigure}

  \begin{subfigure}[b]{\columnwidth}
     \scalebox{1}{\begin{tikzpicture}[line width=0.02cm]

    \node[align=center,minimum size=0.3cm] at (-3.3, 0) {$\mathbf{x}_i$};

    \node[align=center,minimum size=0.5cm] at (-1.3, 0.2) (B1) {\small $\CNN^p(\theta^p)$};
    \node[draw=none,align=center,minimum size=0.5cm] at (-1.3, 1.2) {\small $\CNN^{s}(\theta^s)$};

    \filldraw[fill=green!20!white, draw=green!20!white,rounded corners] (-0.3, -0.3)rectangle(0.3, 0.3);
    \node[draw=none,align=center,minimum size=0.5cm] at (0.0, -0.0) {\small $\mathbf{h}^p_i$};
    \filldraw[fill=yellow!20!white, draw=yellow!20!white,rounded corners] (-0.3, +0.7)rectangle(0.3, 1.3);
    \node[draw=none,align=center,minimum size=0.5cm] at (0.0, +1.0) {\small $\mathbf{h}^s_i$};

    \node[draw=none,align=center,text width=6cm] at (1.9,+0.0) {$y_i$};

    \draw[->,rounded corners] (-2.8, 0.0)--(-0.35, 0.0);
    \draw[->,rounded corners] (-2.8, 0.0)--(-2.5, 0.0)--(-2.5,1.0)--(-0.35,1.0);
    \draw[draw=blue,->,rounded corners] ( 0.35, 0.0)--( 1.5, 0.0);
    \draw[draw=blue,->,rounded corners] ( 0.35, 1.0)--( 1.0, 1.0)--( 1.0, 0.0)--( 1.5, 0.0);

    \draw[draw=red!60!white,dashed,->,rounded corners] (0.0, 1.3)--(0.0, 1.7)--(1.5, 1.7);

    \node[align=center,minimum size=0.5cm] at (0.9, 1.9) {\small $\operatorname{D}^s(\theta^d)$};
    \node[align=center,minimum size=0.5cm] at (1.8, 1.7) {$d_i$};

    \draw[draw=blue,->,rounded corners] (0.0, -0.3)--(0.0, -0.6)--(2.65, -0.6)--(2.65, 1.7)--(2.1, 1.7);

    \node[align=left,minimum size=0.5cm] at (3.2,0.55) {\small $\operatorname{D}^p(\theta^g)$};
    
    \node[align=center,minimum size=0.5cm] at (0.8, 0.3) {\small $\theta^c$};

\end{tikzpicture}

}
     \caption{Domain-generative (\gen) model.}
     \label{fig:cp}
   \end{subfigure}

  \begin{subfigure}[b]{\columnwidth}
     \scalebox{1}{\begin{tikzpicture}[line width=0.02cm]

    \node[align=center,minimum size=0.3cm] at (-3.3, 0) (A2) {$\mathbf{x}_i$};

    \node[align=center,minimum size=0.5cm] at (-1.3, 0.25) (B2) {\small $\CNN (\theta)$};

    \filldraw[fill=green!20!white, draw=green!20!white,rounded corners] (-0.3, -0.3)rectangle(0.3, 0.3);
    \node[draw=none,align=center,minimum size=0.5cm] at (0.0, +0.0) {\small $\mathbf{h}$};
    \node[draw=none,align=center,text width=6cm] at (1.9,+0.0) {$y_i$};

    \draw[->] (-2.8,  0.0)--( -0.35, -0.0);
    \draw[draw=blue,->] ( 0.35, 0.0)--(1.5, 0.0);
    
    \draw[draw=red!60!white,dashed,->,rounded corners] (0., 0.3)--(0., 1.0)--(1.5, 1.0);
    \node[align=center,minimum size=0.5cm] at (0.9, 1.2) {\small $\operatorname{D}^s(\theta^d)$};
    \node[align=center,minimum size=0.5cm] at (1.9, 1.0) {$d_i$};
    \node[align=center,minimum size=0.5cm] at (0.8, 0.3) {\small $\theta^c$};

\end{tikzpicture}

}
     \caption{Baseline \CNN model with adversarial loss.}
     \label{fig:bs}
  \end{subfigure}
  \caption{Proposed model architectures, showing a single training
    instance $(\mathbf{x}_i, y_i)$ with domain
    $d_i$, and baseline model with domain adversarial loss.
    $\operatorname{CNN}$ denotes a convolutional network, 
    $\operatorname{D}$ indicates a discriminator ($d$ for domain adversarial and $g$ for domain generation), and
    \textcolor{red!60!white}{red dashed} and \textcolor{blue}{blue}  lines
    denote adversarial and standard loss, resp.}
    \label{fig:spcp}
\end{figure}
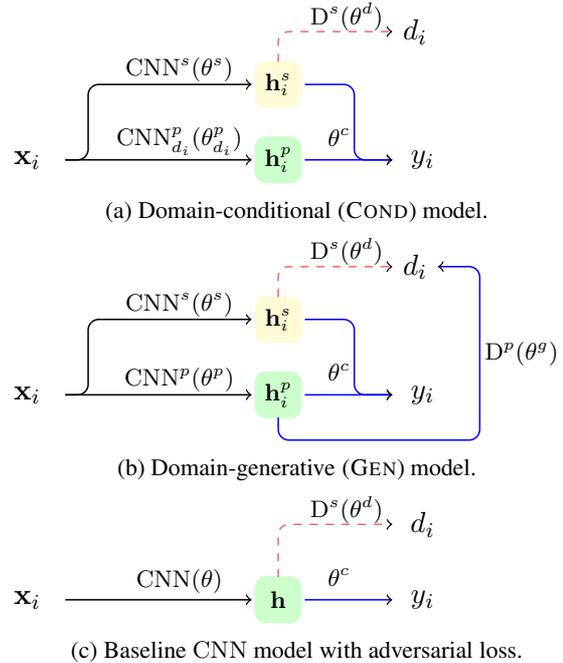

To address these issues, we propose two architectures as illustrated in \figref{fig:spcp} (a and b), parameterised as a convolutional network (CNN) over the input instance, chosen based on the success of CNNs for text categorisation problems \cite{kim2014convolutional}; note, however, that our method is general and can be applied with other network types. 
Both representations are based on the idea of twin representations of each instance,\footnote{This differs from standard architectures, e.g., `baseline' in \figref{fig:bs}, which uses a single representation.} denoted \emph{shared} and \emph{private} representations, which are trained to capture domain-general versus domain-specific concepts, respectively.
This is achieved using various loss functions, most notably an adversarial loss to discourage learning of domain-specific concepts in the shared representations.
The two architectures differ in whether the domain is provided as an input (\cond) or an output (\gen). Below, we elaborate on the details of the two models.

\subsection{Domain-Conditional Model (\cond)}
\label{sec:cond}

The first model, illustrated in \figref{fig:sp}, includes a collection of domain-specific \CNN{}s,  and for each training instance $\xvec$, the domain-specific $\CNN_{d_i}^p$ is used to compute its private representation $\hvec^p$. 
In this manner, the model \emph{conditions} on the domain identifier.
The \cond model also computes a shared representation, $\hvec^s$, directly from $\xvec$, using a shared $\CNN^s$, and the two representations are concatenated together to form input to linear softmax classification function $c$ for predicting class label $y$.
Thus far, the approach resembles \newcite{daume2007frustratingly}, a method for multitask learning based on feature augmentation in a linear model, which works by replicating the input features to create both general shared features, and domain-specific features.  
Note that the approaches differ in that our method uses deep learning to form the two representations, in place of feature replication.

\paragraph{Adversarial Supervision}
A key challenge for the \cond model is that the `shared' representation can be contaminated by domain-specific concepts.
To address this, we borrow ideas from adversarial learning \cite{DBLP:conf/nips/GoodfellowPMXWOCB14,ganin2016domain}.
The central idea is to learn a good general representation (suitable for the shared component) to maximize end task performance, yet obscure the domain information, as modelled by a discriminator, $\operatorname{D}^s$.
This reduces the domain-specific information in the shared representation, however note that important domain-specific components can still be captured in the private representation.

Overall, this results in the training objective:
\begin{align}
\begin{split}
\loss^{\text{\cond}} = \min_{\theta^c,\theta^s,\{\theta^p_\cdot\}} & \max_{\theta^{d}} \xent(\yvec | \hmat^s, \hmat^p, \dvec; \theta^c) \\
&- \lambda_d \xent(\dvec|\hmat^s; \theta^d)
\end{split}
\end{align}
where $\xent$ denotes the cross-entropy classification loss, $\hmat^s = \{ \hvec_i^s(\xvec_i) \}_{i=1}^n$ are the shared representations for the training set of $n$ instances, and likewise  $\hmat^p = \{ \hvec_i^p(\xvec_i, d_i) \}_{i=1}^n$ are the private representations, which are both functions of $\theta^s$ and $\{\theta^p_\cdot\}$, respectively.
Note the negative sign of the adversarial loss (referred to as $d$), and the maximisation with respect to the discriminator parameters $\theta^d$.
This has the effect of learning a maximally accurate discriminator \emph{wrt} $\theta^d$, while making it maximally inaccurate \emph{wrt} representation $\hmat^s$, and is implemented using a gradient reversal step during backpropagation   \cite{ganin2016domain}.

\paragraph{Minimum Entropy Inference}

As \cond conditions on the domain, this imposes the requirement that the domain of the test data is known (and covered in training), which is incompatible with our goal of unsupervised adaptation.
To deal with this situation, we consider each domain in the test set as belonging to one of the training domains, and then select the domain with the minimum entropy classification distribution.
This is based on an assumption that a closely matching domain should be able to make confident predictions.\footnote{The minimum entropy method is quite effective, trailing oracle selection by only $0.8\%$ accuracy.}

\subsection{Domain-Generative Model (\gen)}

The second model is based on \emph{generation} of, rather than \emph{conditioning} on, the domain, which allows the model to learn domain signals that transfer across some, but not all, domains. 
Most components are common with the \cond model as described in  \secref{sec:cond}, including the use of private and shared representations, their use in the classification output, and the adversarial loss based on discriminating the domain from the shared representation.
There are two key differences:
(1) the private representation, $\hvec^p$, is computed using a single $\CNN^p$, rather than several domain-specific \CNN{}s, which confers benefits of domain-generalisation, a more compact model, and simpler test inference;\footnote{The domain need not be known for test examples, so the model can be used directly.} 
and (2) the private representation is used to positively predict the domain, which further encourages the split between domain general and domain-specific aspects of the representation. 

\gen has the following training objective,
\begin{align}
\begin{split}
\loss^{\text{\gen}} & = \min_{\theta^c,\theta^s,\theta^p,\theta^g} \max_{\theta^{d}} \xent(\yvec | \hmat^s, \hmat^p; \theta^c) \\
&~ - \lambda_d \xent(\dvec|\hmat^s; \theta^d) + \lambda_g \xent(\dvec|\hmat^p; \theta^g)
\end{split}
\end{align}
where notation follows that used in \secref{sec:cond}, with the exception of $\hmat^p = \{ \hvec_i^p(\xvec_i) \}_{i=1}^n$ that is redefined, with $\hvec_i^p(\xvec_i)$ a function of $\theta^p$, and the addition of the last term to capture the generation loss $g$.
The same gradient reversal method from \secref{sec:cond} is used during training for the adversarial component.

\section{Experiments}

\subsection{Language Identification}\label{sec:lang_id}
To evaluate our approach, we first consider the language identification task.

\begin{table*}[t!]
  \sisetup{detect-weight=true,detect-inline-weight=math}
  \centering
  \footnotesize
  \begin{tabular}{
    l 
    *{8}{S[table-format=2.1,round-mode=places,round-precision=1]} 
    S[table-format=2.1,round-mode=places,round-precision=1]
    S[table-format=2.1,round-mode=places,round-precision=1]}
    \toprule
    Models  & \dataset{EuroGov} & \dataset{TCL} & \dataset{Wikipedia2} & \dataset{EMEA} & \dataset{EuroPARL} & \dataset{T-BE} & \dataset{T-SC} && $\text{ALL}_{\text{out}}$ \\
    \midrule
    baseline \CNN & 
    \bfseries 99.9 & 91.7 & 88.9 & 93.1 & 98.2 & 85.2 & 92.2  && 92.7  \\
    \qquad $+d$ & 
    \bfseries 99.9 & 92.4 & 88.4 & 90.2 & 98.2 & 87.7 & 93.1  && 92.8  \\
    \qquad $+g$ & 
    \bfseries 99.9 & 92.0 & 88.7 & 91.6 & 98.4 & 86.8 & 92.8  && 92.9  \\
    
    \midrule
    \cond & 
    \bfseries 99.9 & 91.3 & 88.2 & 92.0 & 98.7 & 91.5 & 94.5  && 93.7  \\
    \qquad $+d$ & 
    \bfseries 99.9 & \bfseries  93.5 & 90.1 & 91.3 & 98.7 & 92.6 & \bfseries 97.9  && \bfseries 94.9  \\
    
    \midrule
    \gen & 
    \bfseries 99.9 & 92.3 & 88.0 & 93.3 & 98.6 & 87.1 & 93.8  && 93.3   \\
    \qquad $+d+g$ & 
    \bfseries 99.9 & 93.1 & 88.7 & 92.5 & 99.1 & 91.2 & 96.1  && 94.4   \\

    \midrule
    \resource{langid.py} & 
    98.7 & 90.4 & \bfseries 91.3 & \bfseries 93.4 & \bfseries 99.2 & \bfseries 94.1 & 88.6  && 93.6  \\
    \resource{cld2} & 
    99.0 & 85.0 & 85.3 & 90.7 & 98.5 & 85.0 & 93.4  && 91.0   \\

    \bottomrule
    
  \end{tabular}
  \caption{Accuracy [\%] of the different models over the seven heldout
    datasets, and the macro-averaged accuracy out-of-domain over the 7
    test domains
    (``$\text{ALL}_{\text{out}}$'').
     The best result for each dataset is indicated in \textbf{bold}. Key: $+d$ = domain adversarial, $+g$ = domain generation component.
}
  \label{tab:res}
\end{table*}


\paragraph{Data}
We follow the settings of \newcite{lui2012langid}, involving $5$ training sets from $5$ different domains with $97$ languages in total: \dataset{Debian}, \dataset{JRC-Acquis}, \dataset{Wikipedia}, \dataset{ClueWeb} and \dataset{RCV2}, derived from~\newcite{DBLP:conf/ijcnlp/LuiB11}.\footnote{As \dataset{ClueWeb} in \newcite{lui2012langid} is not publicly accessible, we used a slightly different set of languages but comparable number of documents for training.}
We evaluate accuracy on seven holdout benchmarks:  \dataset{EuroGov},
\dataset{TCL}, \dataset{Wikipedia2}\footnote{Note that the two Wikipedia
  datasets have no overlap.} (all from~\newcite{DBLP:conf/naacl/BaldwinL10}), \dataset{EMEA}~\cite{tiedemann2009news}, \dataset{EuroPARL}~\cite{koehn2005europarl}, \dataset{T-BE}~\cite{tromp2011graph}, and \dataset{T-SC}~\cite{carter2013microblog}.

Documents are tokenized as a byte sequence (consistent with \newcite{lui2012langid}), and truncated or padded to a length of 1k bytes.\footnote{We also tried different document length limits, such as 10k, but observed no substantial change in performance.}

\paragraph{Hyper-parameters}
\label{sec:hyp}

We perform a grid search for the hyper-parameters, and selected the following settings 
to optimise accuracy over heldout data from each of the training domains.
All byte tokens are mapped to byte embeddings, which are random initialized with size $300$.
We use the filter sizes of $2,4,8,16$ and $32$, with $128$ filters for each, to capture $n$-gram features of different lengths.
A dropout rate of $0.5$ was applied to all the representation layers.
We set the factors $\lambda_{d}$ and $\lambda_{g}$ to $10^{-3}$.
All the models are optimized using the Adam Optimizer \cite{kingma2014adam} with a learning rate of $10^{-4}$.

\subsubsection{Results and Analysis}

\paragraph{Baseline and comparisons} For comparison, we implement a \CNN baseline\footnote{The baseline here used a double capacity hidden representation, in order to better match the increased expressivity of the shared/private models.} which is trained using all the data without domain knowledge (i.e.\ the simple union of the different training datasets).
We also employ adversarial learning ($d$) and generation ($g$) of domain to the baseline model to better understand the utility of these methods.
Note that the baseline +$d$ is a multi-domain variant of \newcite{ganin2015unsupervised}, albeit trained without any text in the testing domains.
For our models, we report results of configurations both with and without the $d$ and $g$ components.
We also report the results for two state-of-the-art off-the-shelf LangID tools: (1) \resource{langid.py}\footnote{\url{https://github.com/saffsd/langid.py}} \cite{lui2012langid}; and (2) Google's \resource{cld2}.\footnote{\url{https://github.com/CLD2Owners/cld2}}

\paragraph{Out-of-domain Results}
Our primary concern in terms of evaluating the ability of the different models to generalise, is out-of-domain performance.
\tabref{tab:res} provides a breakdown of out-of-domain results over the $7$ holdout domains.
The accuracy varies greatly between test domains, depending on the mix of languages, length of test documents, etc.
Both our models, \SP and \CP, achieve competitive performance, and are further improved by $d$ and $g$.

For the baseline, applying either $d$ or $g$ results in mild improvements over the baseline, which is surprising as the two
forms of supervision work in opposite directions. 
Overall the small change in performance means neither method appears to be a viable technique for domain adaptation.

Overall, the raw \SP and \CP perform better than the baseline.
Specifically, for \SP, we observed performance gains on \dataset{EuroPARL}, \dataset{T-BE} and \dataset{T-SC}.
These three datasets are notable in containing shorter documents, which benefit the most from shared learning.
However, as discussed earlier, multi-domain data can introduce noise to the shared representation, causing the performance to drop over \dataset{TCL}, \dataset{Wikipedia2} and \dataset{EMEA}.
This observation demonstrates the necessity of applying adversarial learning to \SP.
On the other hand, it is a different story for \CP: vanilla \CP achieves accuracy gains relative to the baseline over $5$ domains, but is slightly below \SP for $4$ domains, a result of parameter-sharing over the private representation.

In terms of the adversarial learning, we see that 
by adding an adversarial component ($+d$ or $+d+g$), \SP and \CP realises substantial improvements out of domain, with the exception of \dataset{EMEA}.
As we motivated, the domain adversarial part $d$ can obscure the domain-specific information in the shared representation, which helps \SP have better generalisation to other domains.
Additionally, applying $g$ to \CP helps the private representation to generalize better.
These results demonstrate that both $d$ and $g$ are necessary components of multi-domain models.
\dataset{EMEA} is noteworthy in that its pattern of results is overall different to the other domains, in that applying $d$ hurts performance.
For this domain, the baseline \CNN performs very well, and \CP does much better than \SP.
We believe the reason is that, as a medical domain, \dataset{EMEA} is very much an outlier and does not align to any single training domain.
Also, there is a lot of borrowing of terms such as drug and disease names verbatim between languages, further complicating the task.

Overall, our best models (\SP$+d$ and \CP$+d+g$) outperform both \resource{Langid.py} and \resource{CLD2} in terms of average out-of-domain accuracy.

\paragraph{In-domain Results}
\tabref{tab:res_id} reports the in-domain performance over the $5$ training domains, using $5$-fold cross validation, as well as the macro-averaged accuracy.
Our proposed methods (\SP$+d$ and \SP$+d+g$) consistently achieve better performance than the baseline.
Both \SP and \CP achieve competitive performance with the state-of-the-art \resource{langid.py} in the in-domain scenario.
Although \resource{langid.py} performs slightly better on average accuracy, our best model outperforms \resource{langid.py} for three of the five datasets.

\begin{table}[t!]
  \sisetup{detect-weight=true,detect-inline-weight=math}
  \centering
  \resizebox{\columnwidth}{!}{%
  \begin{tabular}{l*{1}c*{5}{S[table-format=2.1,round-mode=places,round-precision=1]}c}
    \toprule
    Models & \dataset{Deb} & \dataset{JRCA} & \dataset{Wiki} & \dataset{ClWb} & \dataset{RCV2} & $\text{ALL}_{\text{in}}$ \\
    \midrule
    baseline \CNN &
    96.6  & 99.8  & 97.8  & 90.7  & 97.9   & 96.6 \\
    baseline $+d$ &
    96.5 & 99.8 & 97.8 & 90.7 & 97.0       & 96.4 \\
    
    \midrule
    \SP &
    97.0 & 99.9 & 97.8 & 90.9 & 98.0 &  96.7 \\
    \SP $+d$ &
    97.0 & \bfseries 99.9 & \bfseries 98.4 & 90.8 & 98.1  & 96.8 \\
    
    \CP $+d+g$ &
    \bfseries 97.8 & \bfseries 99.9 & 98.0 & 91.1 & 97.9  & 96.9 \\
    
    \midrule
    \resource{langid.py} &
    97.4 & 99.8 & 97.6 & \bfseries 91.3 & \bfseries 99.3  & \bfseries 97.1 \\
    \resource{cld2} &
    92.2 & 99.8 & 92.3 & 92.3 & 89.8   & 93.3\\

    \bottomrule
    
  \end{tabular}%
  }
  \caption{Accuracy [\%] of different models over five in-domain datasets using cross-validation evaluation and macro-averaged accuracy (``$\text{ALL}_{\text{in}}$'').
  }
  \label{tab:res_id}
\end{table}


\subsection{Product Reviews}

To evaluate the generalization of our methods to other tasks, we experiment with the Multi-Domain Sentiment Dataset \cite{blitzer2007biographies}.\footnote{From \url{https://www.cs.jhu.edu/~mdredze/datasets/sentiment/}, using the positive and negative files from \texttt{unprocessed}, up to 2,000 instances per domain. For the four test domains we automatically aligned the reviews in the \texttt{processed} and \texttt{unprocessed}, such that we can compare results directly against prior work.}
We select the 20 domains with the most review instances, and discard the remaining 5 domains.

For model parameterization, we adopt the same basic hyper-parameter settings and training process as for LangID in \secref{sec:hyp}, but change the filter sizes to $3$, $4$ and $5$, use word-based tokenisation, and truncate sentences to $256$ tokens, for better compatible with shorter documents.

We perform a out-of-domain evaluation over four target domains, ``book'' (\textsf{B}), ``dvd'' (\textsf{D}), ``electronics'' (\textsf{E}) and ``kitchen \& housewares'' (\textsf{K}), as used in \newcite{blitzer2007biographies}. Our experimental setup differs from theirs, in that they train on a single domain and then evaluate on another, while we train over $16$ domains, then evaluate on the four test domains.

\tabref{tab:res_sa_BDEK} presents the results.
Overall, our proposed methods consistently outperform the baselines, with the \CP~$+d+g$ approach a consistent winner over all other techniques.
Note also the lacklustre performance when the baseline is trained with the adversarial loss, mirroring our findings for language identification in \S\ref{sec:lang_id}.
For comparison, we also report the best results of \resource{SCL-MI} and \resource{DANN}, in both cases using an oracle selection of source domain.
Our method consistently outperform these approaches, despite having no test oracle, although note that we use more diverse data sources for training.

\begin{table}[t!]
  \sisetup{detect-weight=true,detect-inline-weight=math}
  \centering
  \footnotesize
  \begin{tabular}{lc*{4}{S[table-format=2.1,round-mode=places,round-precision=1]}}
    \toprule
  Models & & {\textsf{B}} & {\textsf{D}} & {\textsf{E}} & {\textsf{K}} \\
    \midrule
    baseline \CNN & & 79.55 & 81.2 & 86.3 & 87.2
    \\
    baseline $+d$ & & 78.65 & 81.55 & 86.55 & 87.05
    \\
    
    \midrule
    \SP           & & 79.15 & 81.75 & 85.8 & 87.15
    \\ 

    \SP $+d$      & & 79.8 & 82.3 & 86.75 & 87.35
    \\
    
    \CP $+d+g$    & & \bfseries 80.2 & \bfseries 82.35  & \bfseries 87.25 & \bfseries 87.8
    \\
    \midrule
    \resource{SCL\_MI}$^\clubsuit$
                  & & 76.0 & 78.5 & 77.9 & 85.9 \\
    \resource{DANN}$^\diamondsuit$
                  & & 72.3 & 78.4 & 84.3 & 85.4 \\
    \midrule
    \resource{in domain}$^\clubsuit$ 
                  & & 82.4 & 80.4 & 84.4 & 87.7 \\

    \bottomrule
    
  \end{tabular}
  \caption{Accuracy [\%] of different models over $4$ domains (\texttt{B}, \texttt{D}, \texttt{E} and \texttt{K}) under out-of-domain evaluations on Multi Domain Sentiment Dataset.
Key: $^\clubsuit$ from \newcite{blitzer2007biographies}; $^\diamondsuit$ from \newcite{ganin2015unsupervised}.
}
  \label{tab:res_sa_BDEK}
\end{table}


\section{Conclusions}

We have proposed a novel deep learning method for multi-domain learning, based on joint learning of domain-specific and domain-general components, using either domain conditioning or domain generation.
Based on our evaluation over multi-domain language identification and multi-domain sentiment analysis, we show our models to substantially outperform a baseline deep learning method, and set a new benchmark for state-of-the-art cross-domain LangID.
Our approach has potential to benefit other NLP applications involving multi-domain data.

\section*{Acknowledgments}
We thank the anonymous reviewers for their helpful feedback and suggestions, and the National Computational Infrastructure Australia for computation resources.
This work was supported by the Australian Research Council (FT130101105).

\bibliography{ref}
\bibliographystyle{acl_natbib}

\appendix

\end{document}